\documentclass[conference]{IEEEtran}
\IEEEoverridecommandlockouts

\usepackage{cite}
\usepackage{amsmath,amssymb,amsfonts}
\usepackage{graphicx}
\usepackage{textcomp}
\usepackage{xcolor}
\usepackage{tabularx, arydshln, float, booktabs}
\usepackage{algorithm}
\usepackage{algpseudocode}
\usepackage{amsmath}
\usepackage{subcaption}
\usepackage{xcolor}
\usepackage{threeparttable}
\def\BibTeX{{\rm B\kern-.05em{\sc i\kern-.025em b}\kern-.08em
    T\kern-.1667em\lower.7ex\hbox{E}\kern-.125emX}}
\begin{document}

\title{Deep Convolutional Neural Networks Structured Pruning via Gravity Regularization}

\author{\IEEEauthorblockN{Abdesselam Ferdi}
\IEEEauthorblockA{\textit{Electronics Department} \\
\textit{Constantine 1 - Frères Mentouri University}\\
Constantine, Algeria \\
abdesselam.ferdi@gmail.com}
}

\maketitle

\begin{abstract}
Structured pruning is a widely employed strategy for accelerating deep convolutional neural networks (DCNNs). However, existing methods often necessitate modifications to the original architectures, involve complex implementations, and require lengthy fine-tuning stages. To address these challenges, we propose a novel physics-inspired approach that integrates the concept of gravity into the training stage of DCNNs. In this approach, the gravity is directly proportional to the product of the masses of the convolution filter and the attracting filter, and inversely proportional to the square of the distance between them. We applied this force to the convolution filters, either drawing filters closer to the attracting filter (experiencing weaker gravity) toward non-zero weights or pulling filters farther away (subject to stronger gravity) toward zero weights. As a result, filters experiencing stronger gravity have their weights reduced to zero, enabling their removal, while filters under weaker gravity retain significant weights and preserve important information. Our method simultaneously optimizes the filter weights and ranks their importance, eliminating the need for complex implementations or extensive fine-tuning. We validated the proposed approach on popular DCNN architectures using the CIFAR dataset, achieving competitive results compared to existing methods.
\end{abstract}


\section{Introduction}
Deep convolutional neural networks (DCNNs) have demonstrated remarkable proficiency in addressing a variety of computer vision tasks. Despite this, a significant portion of the research community remains primarily concentrated on enhancing the accuracy of DCNNs, often overlooking critical factors such as computational efficiency and memory requirements. More specifically, the energy consumption, quantified by the number of floating point operations per second (FLOPs), and the memory footprint, determined by the number of parameters, are frequently disregarded during the performance optimization process.

DCNNs are predominantly deployed on resource-constrained devices, such as smartphones. However, early ImageNet models, such as VGG-19 and ResNet-50, are characterized by their high complexity, with approximately $143.7$M and $25.6$M parameters, and computational requirements of $0.8$ and $4.09$ GFLOPS, respectively. This complexity poses a significant challenge in scenarios where efficiency is paramount. Consequently, there is a growing demand for the development of lightweight models that maintain a balance between compactness, in terms of power and memory, and performance, with minimal accuracy degradation. The question then arises: how can we effectively transform these heavyweight models into efficient, lightweight counterparts?

Numerous strategies have been proposed to answer this question, with model compression emerging as the most prevalent approach \cite{zawish2024complexity}. This approach encompasses a variety of techniques, including quantization, low-rank approximation, distillation, and pruning methods. Pruning can be further categorized into semi-structured, unstructured, and structured approaches. Recent research has primarily concentrated on structured pruning (SP), which forms the basis of our work \cite{ferdi2024electrostaticforceregularizationneural,gupta2024torque}. In SP, entire filters or channels are removed from the convolutional layers of the model.

This work addresses the following critical question: What is the optimal way to configure a model for the pruning stage? 

We put forth a novel, physics-inspired method for the SP of DCNNs. In particular, we incorporate the concept of gravity into the training stage of DCNNs. This concept is leveraged during model training to redistribute the weights of the filters around a designated attracting filter within the convolutional layer. The term ’attracting filter’ refers to the convolution filter with the largest mass, determined by the L$_1$-norm, or the filter located at the zero position within the convolutional layer. Essentially, we induced sparsity in the convolution filters, which is directly proportional to the product of the masses of the convolution filter and the attracting filter, and inversely proportional to the square of the distance between them. Upon completion of the model training, we can prune the sparser filters based on a predetermined pruning ratio. Our method effectively reduces the model’s complexity while preserving its accuracy. Moreover, it does not necessitate any modifications to the original model architecture, and allows for the compression of the trained model at varying pruning ratios without the need for retraining. This latter attribute distinguishes our method from existing SP techniques, which typically train the model for a specific pruning ratio and necessitate retraining when this ratio is altered. 

Our contributions are as follows: 
\begin{itemize}
    \item We introduce a novel, physics-inspired SP method for DCNNs, grounded in the concept of gravity.
    \item Our method effectively reduces the complexity of DCNNs, both in terms of energy consumption and memory requirements.
    \item The proposed method is easy to implement and does not necessitate any modifications to the original architecture. It can be applied to both non-trained and pre-trained models.
    \item We can prune a model trained with our gravity-based method at varying pruning ratios without the need for retraining.
\end{itemize} 

The rest of the paper is organized as follows. Section 2 reviews existing methods for pruning DCNNs. In Section 3, we introduce our gravity methodology, including its motivation and the proposed approach. Section 4 details the experimental setup, followed by the presentation of results in Section 5. Section 6 provides a discussion of these results, while Section 7 summarizes the work and explores potential future directions.

\section{Related Work}
The development of complex deep models continues to deliver promising results, making model acceleration a key research area. In the past decade, model pruning has gained significant attention due to the demand for efficient deep networks. Pruning is categorized based on its timing relative to network training: at initialization, during training, and after training. These categories include semi-structured \cite{Grimaldi_2023_ICCV}, unstructured \cite{zhang2022advancing}, and SP \cite{ferdi2024electrostaticforceregularizationneural}. Our focus is on SP after training, which involves pruning a densely pretrained model and subsequently fine-tuning it to recover any performance loss caused by pruning \cite{gupta2024torque,ferdi2024electrostaticforceregularizationneural}. SP techniques aim to eliminate redundant connections with minimal impact on overall performance, thereby improving test-time efficiency. However, they often significantly increase training time, as the extensive fine-tuning required can nearly double the overall training duration.

\section{Methodology}
This section commences with a discussion of the motivation for incorporating the concept of gravity from physics to the SP of DCNNs, and proceeds with a thorough explanation of the proposed method.

\subsection{Motivation}
In the realm of physics, gravity, or gravitational force, is a ubiquitous natural phenomenon. This force manifests between objects possessing mass, drawing them toward each other. The intensity of gravity is contingent upon two factors: mass and distance. The larger the masses of the objects, the stronger the gravitational pull between them. Conversely, the greater the distance between the objects, the weaker the gravitational force.

We posit that the introduction of such a force into the training of DCNNs will facilitate the redistribution of the weights of convolution filters around a specified location. By exerting a gravitational force on the filters of a convolutional layer during the training stage of DCNNs, we induce an increasing rate of sparsity in the weights of the filters. The further a filter is from a specified location (i.e., the attracting filter), the greater the induced sparsity. 
    
\subsection{Proposed Method}
In physics, gravity is a force that causes objects with mass $m$ to be attracted to one another. Newton's law of gravitation, denoted as $F$, is expressed as
\begin{align}
  F = G \frac{m_1 m_n}{d^2} 
  \label{eq1}
\end{align}
where $G$ represents the gravitational constant ($6.7\times10^{-11} N.m^2.kg^{-2}$). $m_1$ and $m_n$ denote the masses of two objects, and $d$ is the distance between them. 

Analogous to Newton's law of gravitation, we define the gravitational force between two filters in a DCNN using Eq. \ref{eq1}. Here, $m_1$ and $m_n$ represent the masses corresponding to the attracting filter and the convolution filter $n$, respectively.  The gravitational constant is denoted by $G$, and $r$ represents the distance between the attracting filter and the convolution filter $n$.

To understand the gravity, consider a convolutional layer, denoted as $l$, consisting of $N$ filters. Each filter has dimensions of $[C, K, K]$, where $C$ represents the number of channels and $K$ denotes the filter size. At layer $l$, the attracting filter generates a gravitational field that exerts the following effects: (\textbf{1}) an attractive force on neighboring filters, and (\textbf{2}) no force on itself.

We define the mass of filter $n$ in layer $l$ as the L$_1$-norm of its weights, thus
\begin{align}
    m_n = \|W_{n,l}\|_1
    \label{eq2}
\end{align}

where $n \in [0, N-1]$.

The position of filter $n$ (or $p_n$) does not correspond to a physical position in layer $l$ as the filters are not physically positioned in a specific way. However, we can refer to the physical position of filter $n$ by its index in layer $l$. For example, the first filter in layer $l$ could be said to be at position zero, the second filter at position one, and so on. Based on the given definition, the distance between the attracting filter and the convolution filter $n$ in layer $l$ is determined by computing the absolute difference between their respective indices. Thus, we can write
\begin{align}
  d = |p_1 - p_n|
  \label{eq3}
\end{align}

where $p_1$ and $p_n$ represent the position of the attracting filter and the convolution filter $n$ in layer $l$, respectively. 

In DCNNs, the influence of the gravitational force is reflected in the weights of the convolution filters. Specifically, filters subjected to stronger forces will have weights that converge to zero, whereas filters experiencing weaker forces will retain non-zero weights. 

Our objective is to make filters located in proximity to the attracting filter (i.e., at small distances) maintain non-zero weights, corresponding to weaker gravitational forces. Conversely, filters positioned farther from the attracting filter (i.e., at greater distances) should have zero weights, corresponding to stronger gravitational forces. However, as indicated by Eq. \ref{eq1}, the gravitational force $F$ is inversely proportional to the distance $d$, resulting in weaker forces at greater distances and stronger forces at shorter distances. 

To align with our objective of ensuring that the gravitational force $F$ becomes weaker for filters in close proximity to the attracting filter and stronger for those farther away, we reformulate the distance $d$ (Eq. \ref{eq2}) as follows

\begin{align}
  d = \frac{1}{|p_1 - p_n|}
  \label{eq4}
\end{align}

We generalize the Eq. \ref{eq1} for the $n^{th}$ filter in layer $l$ as
\begin{align}
    F_{n,l} = G\frac{m_{1, l}m_{n,l}}{d_{n,l}^2} 
    \label{eq5}
\end{align}

Here, it is important to emphasize that the gravitational force becomes nonexistent when considering the attracting filter ($m_{1,l}$) interacts with itself.

we incorporate the gravitation force to regularize the cost function
\begin{align}
    \Tilde{J} = \sum_{n,l} J(w_{n,l}; X, y) + \alpha_g F_{n,l}
    \label{eq6}
\end{align}


Here, $\alpha_g$, referred to as the gravity rate, is a hyperparameter that adjusts the relative contribution of the penalty term $F_{n,l}$ relative to the standard objective function $J$.

\begin{figure}[h]
    \includegraphics[height=0.4\textwidth, width=\linewidth]{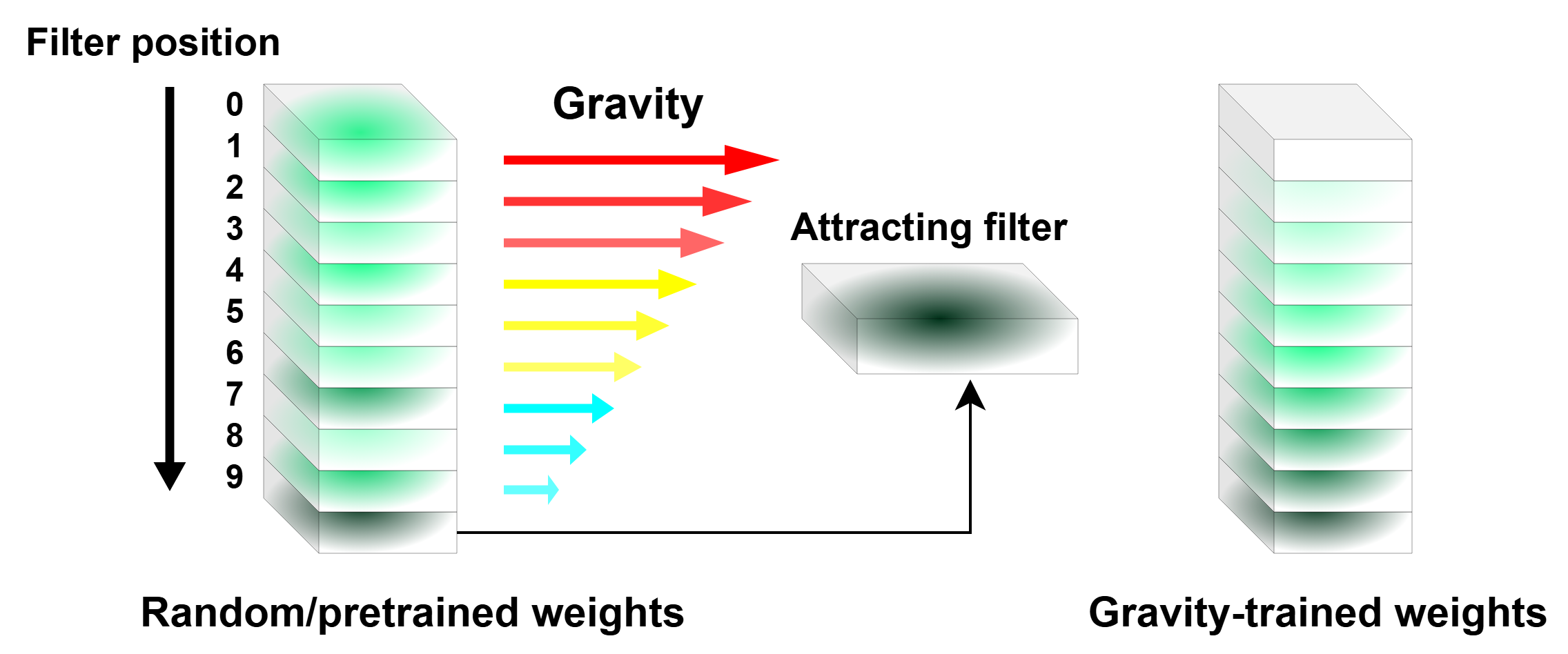}
    \caption{An illustration of gravity-based training: A convolutional layer comprising ten filters is presented as an example. The color shading indicates the mass of each filter, with the attracting filter assigned the largest mass. Initially, the convolutional layer contains filters initialized with either random or pretrained weights. Filters located farther from the attracting filter experience stronger gravitational forces, driving their weights toward zero. In contrast, filters in closer proximity encounter weaker forces, thereby pulling their weights toward non-zero values.}
  \label{figure1}
\end{figure}

When the optimizer minimizes the regularized objective function $\Tilde{J}$, it simultaneously reduces both the original objective function $J$ on the training data and the gravitational force $F_{n,l}$. By minimizing $F_{n,l}$, the optimizer drives the weights of filters experiencing larger gravitational forces toward zero. The influence of the gravitational force on filter weights is illustrated in Figure \ref{figure1}.

The gradient of the regularized objective function with respect to the weights is expressed as
\begin{equation}
    \nabla \Tilde{J} = \sum_{n,l} \nabla_{w_{n,l}} J(w_{n,l}; X, y) + \alpha_g \nabla_{w_{n,l}} F_{n,l}   
    \label{eq7}
\end{equation}

By substituting Eq. \ref{eq5} into Eq. \ref{eq7} (omitting indices $n$ and $l$), we can write
\begin{align}
    \nabla \Tilde{J} = \nabla_w J(w; X, y) + \alpha_g \nabla_w G \frac{m_1 m_n}{d^2} 
    \label{eq8}
\end{align}

\begin{align}
    \nabla \Tilde{J} = \nabla_w J(w; X, y) + \alpha_g G \frac{m_1}{d^2}(w)
    \label{eq9}
\end{align}

where the function (.) is the sign function.

The weights are updated through a single gradient step, using the following learning rule
\begin{align}
    w \leftarrow w - \epsilon(\nabla_w J(w; X, y) + \alpha_g  G \frac{m_1}{d^2}(w))
    \label{eq10}
\end{align}

where $\epsilon$ is the learning rate.

By examining Eq. \ref{eq10}, we can see that the contribution of the gravitational force to the gradient is a variable factor with a sign corresponds to the sign of the weight of the filter $n$ (i.e., sign($w$)). This factor is dependent on the distance $d$. Consequently, filters that are farther from the attracting filter ($m_1$) will have a larger gradient value, while filters closer to the attracting filter will exhibit a smaller gradient value.

To facilitate understanding, the proposed method is outlined in Algorithm \ref{algorithm1}. Notably, the gravitational force was only applied to the convolutional layers to prune. This approach is justified as follows: during the optimization process, minimizing the regularized objective function $\Tilde{J}$ simultaneously reduces both the standard objective function $J$ and the gravitational force $F$. The reduction of the standard objective function directs the filter weights toward local minima, while minimizing the gravity adjusts the weights of the filters to prune toward a specific weight distribution. This adjustment ensures that the removal of these filters results in low information loss.

\begin{algorithm}[t]
\caption{Gravity-based training algorithm}
\label{algorithm1}
\begin{algorithmic}[1]
\State \textbf{Input:} Randomly initialized or pretrained model $M_{baseline}$, convolutional layers to prune $L$, gravitational constant $G$, gravity rate $\alpha_g$, epochs $E$
\State \textbf{Output:} Gravity-trained model $M_{Gravity}$
\For{layer $l$ in $L$}
    \State Extract and flatten the weights of filters of layer $l$
    \For{filter $n$ in layer $l$}
        \State Compute the mass of the convolution filter $n$
        \State Rank the masses of filters and determine the attracting filter $m_{1,l}$ 
        \State Compute the distances $d_{n,l}$ between the convolution filter $n$ and the attracting filter
        \State $F_{n,l} = G \frac{|m_{1,l}| |m_{n,l}|}{d_{n,l}^2}$
    \EndFor
\EndFor
\State Compute the sum of the gravitational forces ($F$)
\State Regularize the standard objective function $J$ based on $\alpha_g$
\State Train the model for $E$ epochs using SGD with the regularized objective function $\Tilde{J}$
\end{algorithmic}
\end{algorithm}

\section{Experimental Setup}
In this section, we present the details of the experiments, including the dataset utilized, the models employed, and the implementation of the proposed gravity-based approach.

\subsection{Datasets and Networks}
We conducted analyses on the CIFAR dataset \cite{ferdi2024electrostaticforceregularizationneural} with ResNet \cite{ferdi2024electrostaticforceregularizationneural} and VGGNet \cite{ferdi2024electrostaticforceregularizationneural}. The model's parameters were optimized using the stochastic gradient descent (SGD) optimizer, implemented with PyTorch 2.4.1 on an NVIDIA Tesla T4 GPU provided by Google Colaboratory \cite{googlecolab}. 

\subsection{Pruning and Fine-tuning}
After completing the gravity-based model training, we move to the pruning phase, where filters with zero or less important weights are removed to minimize information loss. A local pruning strategy is used, ranking filters by their L$_1$-norm and eliminating those with the smallest norm based on a consistent pruning rate across convolutional layers. We summarized the pruning strategy in Algorithm \ref{algorithm2}.

After the pruning stage, for fair comparison with existing SP methods, the pruned models were fine-tuned using the same hyperparameters (e.g., epochs, learning rate, batch size) as other pruning methods \cite{wang2021neural}.

\begin{algorithm}[b]
\caption{Local structured pruning algorithm}
\label{algorithm2}
\begin{algorithmic}[1]
    \State \textbf{Input:} Gravity-trained model $M_{Gravity}$, convolutional layers to prune $L$, pruning ratio $p_{r}$
    \State \textbf{Output:} Pruned model $M_{pruned}$
    \For{layer $l$ in $L$}  
        \State Extract and flatten the weights of filters of layer $l$
        \For{filter $n$ in layer $l$} 
            \State Compute the L$_1$-norm of filter $n$
        \EndFor
        \State Rank the filters based on their L$_1$-norms
        \State Prune the filters with lowest L$_1$-norms based on the pruning rate $p_{r}$ 
    \EndFor
\end{algorithmic}
\end{algorithm}

\section{Results}
In this section, we present the results of the models trained using the gravity method, both prior to and following fine-tuning. Furthermore, we present the training cost associated with the method.

\subsection{Pruning Results}
\begin{figure*}[h]
  \centering
  \includegraphics[width=\linewidth]{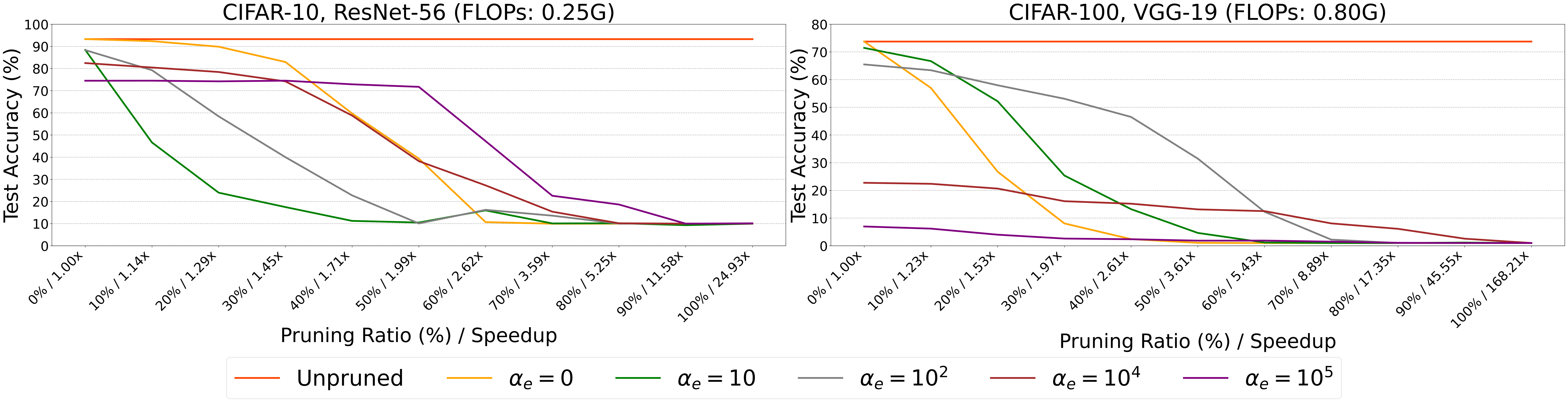}
  \caption{Top-1 accuracy of pruned ResNet-56 and VGG-19 models, initialized with pretrained weights and trained with gravity at five distinct gravity rates, without any fine-tuning.}
  \label{figure2}
\end{figure*}

The performance of the CIFAR-pretrained ResNet-56 and VGG-19 models, trained with the gravity method at gravity rates of $10^5$ and $10^2$, respectively, prior to fine-tuning, is reported in Table \ref{table1}. The baseline networks were trained on the same CIFAR dataset, achieving accuracies comparable to those reported in the original publications. Additionally, the results at different gravity rates are presented as curves in Figure \ref{figure2}. In this figure, for each network, the pruned top-1 accuracy is plotted against the pruning ratio (along with the corresponding speedup rate), which ranges from $0\%$ to $100\%$.

\begin{table}[b]
\begin{threeparttable}
  \caption{Pruning results of the ResNet-56/CIFAR-10 and VGG-19/CIFAR-100 models}
  \centering
    \begin{tabularx}{\columnwidth}{>{\raggedright\arraybackslash}p{2cm}>{\centering\arraybackslash}X>{\centering\arraybackslash}X>{\centering\arraybackslash}X>{\centering\arraybackslash}X>{\centering\arraybackslash}X}
    \toprule
    \multicolumn{6}{l}{\textbf{ResNet-56 + CIFAR-10:} {\scriptsize Baseline Acc. 93.39\%, Params. 0.85M, FLOPs 0.25G}} \\
    \hline
    Pruning rate (\%) & $10$ & $20$ & $30$ & $40$ & $50$ \\
    \hline
    Speedup & $1.14\times$ & $1.29\times$ & $1.45\times$ & $1.71\times$ & $1.99\times$ \\
    \hline
    Compression & $1.13\times$ & $1.26\times$ & $1.45\times$ & $1.69\times$ & $2.00\times$ \\
    \hline
    Pruned Acc. (\%) & $74.52$ & $74.22$ & $74.53$ & $72.88$ & $71.74$ \\
    \toprule
    \multicolumn{6}{l}{\textbf{VGG-19 + CIFAR-100:} {\scriptsize Baseline Acc. 73.76\%, Params. 20.07M, FLOPs 0.80G}} \\
    \hline
    Pruning rate (\%) & $10$ & $20$ & $30$ & $40$ & $50$ \\
    \hline
    Speedup & $1.23\times$ & $1.53\times$ & $1.97\times$ & $2.61\times$ & $3.61\times$ \\
    \hline
    Compression & $1.24\times$ & $1.57\times$ & $2.04\times$ & $2.78\times$ & $3.98\times$ \\
    \hline
    Pruned Acc. (\%) & $63.39$ & $57.96$ & $53.10$ & $46.52$ & $31.47$ \\
    \bottomrule
    \end{tabularx}
  \label{table1}
\end{threeparttable}
\end{table}

\subsection{Training Overhead}
The training times without fine-tuning of the ResNet-56 and VGG-19 models on the CIFAR dataset are presented in Table \ref{table2}.

\begin{table}[t]
    \begin{threeparttable}
  \caption{Comparison of training costs for ResNet-56 and VGG-19 models trained using the baseline method, L$_1$-norm regularization, and the proposed gravity method on an NVIDIA Tesla T4 GPU.}
  \centering
    \begin{tabularx}{\columnwidth}{>{\centering\arraybackslash}X>{\centering\arraybackslash}X>{\centering\arraybackslash}X}
    \toprule
    Model/Dataset & Method & Training time (h)\\ 
    \hline
    & Baseline & $2.84$ \\
    ResNet-56/CIFAR-10 & L$_1$-norm & $4.15$ \\
    & Gravity (\textbf{ours}) & $5.10$ \\
    \hline
    & Baseline & $2.65$ \\
    VGG-19/CIFAR-100 & L$_1$-norm & $13.08$ \\
    & Gravity (\textbf{ours}) & $16.94$ \\
    \bottomrule    
    \end{tabularx}
    \label{table2}
    \end{threeparttable}
\end{table}

\subsection{Fine-tuning Results}
We evaluated our gravity method on the CIFAR dataset using ResNet and VGGNet, as shown in Tables \ref{table3} and \ref{table4}. For existing SP methods, we used the original reported results. 

\begin{table}[t]
\begin{threeparttable}
  \caption{Comparison of different methods on CIFAR-10 with ResNet-56}
  \centering
    \begin{tabular}{lllll}
    \toprule
    Method & Base & Pruned & Acc. & Speed \\ 
    & Acc. (\%) & Acc. (\%) & drop & up \\
    \hline
    GReg-1 \cite{wang2021neural} & $93.51$ & $93.25$ & $0.26$ & $1.99\times$ \\
    GReg-2 \cite{wang2021neural} & $93.51$ & $93.28$ & $0.23$ & $1.99\times$ \\ 
    CP \cite{he2017channel} & $92.80$ & $91.80$ & $1.00$ & $2.00\times$ \\
    AMC \cite{he2018amc} & $92.80$ & $91.90$ & $0.90$ & $2.00\times$ \\
    SFP \cite{he2018soft} & $93.59$ & $93.36$ & $0.23$ & $2.11\times$ \\
    WHC \cite{chen2023whc} & $93.59$ & $93.74$ & $0.15$ & $2.11\times$ \\
    LFPC \cite{he2020learning} & $93.59$ & $93.24$ & $0.35$ & $2.12\times$ \\
    Torque \cite{gupta2024torque} & $93.48$ & $93.76$ & -$0.28$ & $2.15\times$ \\
    Electrostatique force \cite{ferdi2024electrostaticforceregularizationneural} & $\mathbf{94.05}$ & $\mathbf{93.88}$ & $0.17$ & $\mathbf{2.17\times}$ \\
    Gravity (r) (\textbf{ours}) & $88.02$ & $91.07$ & -$\mathbf{3.05}$ & $\mathbf{2.17\times}$ \\
    Gravity (p) (\textbf{ours}) & $93.53$ & $92.99$ & $0.54$ & $\mathbf{2.17\times}$ \\
    \hline
    ABC Pruner \cite{lin2020channel} & $93.26$ & $93.23$ & $0.03$ & $2.18\times$ \\
    RL-MCTS \cite{wang2022channel} & $93.20$ & $93.56$ & -$0.36$ & $2.22\times$ \\
    C-SGD \cite{ding2019centripetal} & $93.39$ & $93.44$ & -$0.05$ & $2.55\times$ \\ 
    GReg-1 \cite{wang2021neural} & $93.51$ & $93.18$ & $0.18$ & $2.55\times$ \\
    GReg-2 \cite{wang2021neural} & $93.51$ & $93.36$ & $0.00$ & $2.55\times$ \\
    AFP \cite{ding2018auto} & $93.93$ & $92.94$ & $0.99$ & $2.56\times$ \\
    Torque \cite{gupta2024torque} & $93.48$ & $93.40$ & $0.08$ & $2.60\times$ \\
    Electrostatique force \cite{ferdi2024electrostaticforceregularizationneural} & $\mathbf{94.05}$ & $\mathbf{93.64}$ & $0.41$ & $\mathbf{2.62\times}$ \\
    Gravity (r) (\textbf{ours}) & $88.02$ & $90.21$ & -$\mathbf{2.19}$ & $\mathbf{2.62\times}$ \\
    Gravity (p) (\textbf{ours}) & $93.53$ & $92.73$ & $0.8$ & $\mathbf{2.62\times}$ \\
    \hline
    WHC \cite{chen2023whc} & $93.59$ & $93.29$ & $0.30$ & $2.71\times$ \\
    Torque \cite{gupta2024torque} & $93.48$ & $93.26$ & $0.22$ & $2.72\times$ \\
    Electrostatique force \cite{ferdi2024electrostaticforceregularizationneural} & $\mathbf{94.05}$ & $\mathbf{93.57}$ & $0.48$ & $\mathbf{2.73\times}$ \\
    Gravity (r) (\textbf{ours}) & $88.02$ & $90.43$ & -$\mathbf{2.41}$ & $\mathbf{2.73\times}$ \\
    Gravity (p) (\textbf{ours}) & $93.53$ & $92.67$ & $0.86$ & $\mathbf{2.73\times}$ \\
    \bottomrule
  \end{tabular}
  \label{table3}
    \begin{tablenotes}
        \footnotesize
        \item A negative value in '\upshape Acc. drop' indicates an improved model accuracy. 
        \item '\upshape r' and '\upshape p' denote gravity-trained models with randomly initialized weights and pretrained weights, respectively.
    \end{tablenotes}
\end{threeparttable}
\end{table}

\begin{table}[t]
\begin{threeparttable}
  \caption{Comparison of different methods on CIFAR-100 with VGG-19}
  \centering
    \begin{tabular}{lllll}
    \toprule
    Method & Base & Pruned & Acc. & Speed \\ 
    & Acc. (\%) & Acc. (\%) & drop & up \\
    \hline
    Kron-OBD \cite{wang2019eigendamage} & $73.34$ & $60.70$ & $12.64$ & $5.73\times$ \\
    Kron-OBS \cite{wang2019eigendamage} & $73.34$ & $60.66$ & $12.68$ & $6.09\times$ \\
    Electrostatique force \cite{ferdi2024electrostaticforceregularizationneural} & $\mathbf{74.59}$ & $\mathbf{69.00}$ & $5.59$ & $\mathbf{6.85\times}$ \\
    Gravity (r) (\textbf{ours}) & $58.62$ & $61.96$ & -$\mathbf{3.34}$ & $\mathbf{6.85\times}$ \\
    Gravity (p) (\textbf{ours}) & $70.99$ & $68.84$ & $2.15$ & $\mathbf{6.85\times}$ \\
    \hline
    EigenDamage \cite{wang2019eigendamage} & $73.34$ & $65.18$ & $8.16$ & $8.80\times$ \\
    GReg-1 \cite{wang2021neural} & $74.02$ & $67.55$ & $6.47$ & $8.84\times$ \\
    GReg-2 \cite{wang2021neural} & $74.02$ & $\mathbf{67.75}$ & $6.27$ & $8.84\times$ \\
    Torque \cite{gupta2024torque} & $73.03$ & $65.87$ & $7.16$ & $8.88\times$ \\
     Electrostatique force \cite{ferdi2024electrostaticforceregularizationneural} & $\mathbf{74.59}$ & $67.53$ & $7.06$ & $\mathbf{8.89\times}$ \\
    Gravity (r) (\textbf{ours}) & $58.62$ & $60.70$ & -$\mathbf{2.08}$ & $\mathbf{8.89\times}$ \\
    Gravity (p) (\textbf{ours}) & $70.99$ & $66.79$ & $4.2$ & $\mathbf{8.89\times}$ \\
    \bottomrule
  \end{tabular}
  \label{table4}
\end{threeparttable}
\end{table}

\section{Discussion}
In this section, we first discuss the pruning results of the gravity-trained networks and the associated training overhead. Subsequently, we compare our findings to state-of-the-art (SOTA) SP methods and provide a theoretical analysis of the results. Finally, we perform an ablation study to investigate the impact of the gravity rate on model's performance.

\subsection{Gravity-based Pruning}
Table \ref{table1} shows that a network trained using our gravity method can be pruned by up to $40\%$, resulting in a reduction of the baseline accuracy by approximately $37\%$. For example, a ResNet-56 model trained on the CIFAR-10 dataset using the gravity method experiences a $23\%$ reduction in baseline accuracy when pruned at a rate of $50\%$.

In terms of training time, Table \ref{table2} demonstrates that the baseline method achieved the shortest training time for both ResNet-56 and VGG-19. In contrast, the L$_1$-norm and gravity methods resulted in increased training durations. For example, the ResNet-56 model trained on the CIFAR-10 dataset using the baseline method completed in $2.84$ hours. In comparison, the L$_1$-norm method extended the training time by approximately $46\%$, requiring $4.15$ hours. The proposed gravity method incurred the highest training cost, with a total training time of $5.10$ hours, representing an $80\%$ increase relative to the baseline. This increase in training time for the gravity method is due to the additional computational overhead introduced by the calculation of the penalty term incorporated into the gradient (Eq. \ref{eq10}) during the optimization process.

Overall, while the gravity method imposes the highest training cost compared to the baseline and standard L$_1$-norm methods, its demonstrated effectiveness in enhancing the performance of pruned models (as illustrated in Figure \ref{figure2}) may justify the additional computational expense. This is particularly relevant in high-stakes applications, such as autonomous driving and healthcare  \cite{ferdi2024quadratic, ferdi2024residual}), or in scenarios with ample computing resources.

\subsection{Comparison with State-of-the-arts}
Our Gravity (r) approach, while achieving higher speedups, consistently results in minimal accuracy drop compared to all existing methods. In terms of top-1 accuracy post-pruning, we observe the following: \textbf{(1)} ResNet-56/CIFAR-10: our Gravity (p) method outperforms CP by $1.19\%$ and AMC by $1.09\%$ at a $2.17\times$ speedup. Although Torque demonstrates a $0.77\%$ advantage at the same speedup, this margin diminishes at higher speedups. At a $2.73\times$ speedup, WHC surpasses both Torque and our Gravity (p) method by $0.03\%$ and $0.62\%$, respectively. However, Electrostatique force outperforms WHC by $0.28\%$ at the same speedup. \textbf{(2)} VGG-19/CIFAR-100: Our Gravity (p) method exceeds Kron-OBD by $8.14\%$ at a $6.85\times$ speedup. At an $8.89\times$ speedup, it outperforms EigenDamage by $1.61\%$ and Torque by $0.92\%$. Nonetheless, the Electrostatic Force method outperforms our Gravity (p) by $0.74\%$ at the same speedup. Although GRreg-2 offers a marginal $0.22\%$ improvement, it incurs substantially higher computational costs. Furthermore, GReg-1/2 necessitates full retraining for any changes in pruning rate, whereas our approach remains adaptive and cost-efficient without requiring complete retraining.

\subsection{Theoretical Analysis}
To theoretically interpret the obtained results, we refer to Eq. \ref{eq10}. The penalty term introduced into the gradient arises from the gravitational force. When the distance $d$ is larger, the penalty term decreases, allowing the weights to retain non-zero values. Conversely, when $d$ is smaller, the penalty term increases, causing the weights to approach zero. By pruning filters with zero weights, the model can be accelerated while retaining critical information from filters with non-zero weights. This analysis supports the conclusion that our gravity-based training method is effective for optimally configuring both modern deep networks with residual connections and single-branch architectures for the pruning stage, with minimal accuracy drop.

\subsection{Ablation Study}
Our method relies significantly on the gravity rate, denoted as $\alpha_g$, which serves as a key hyperparameter. This parameter governs the strength of the gravitational force applied to the filters within the convolutional layers of the model, as defined in Eq. \ref{eq6}. A large $\alpha_g$ imposes a substantial penalty on the gradient, leading to pronounced variations in the updated weights. In contrast, smaller values of $\alpha_g$ result in a minimal gradient penalty, corresponding to more moderate weight updates.
  
Selecting an appropriate value for $\alpha_g$ is critical, as it enables a significant proportion of filters to achieve zero weights, thereby facilitating higher pruning ratios without degrading model accuracy. To identify the appropriate value of $\alpha_g$ , we trained models using four different values: $10$, $10^{2}$, $10^{4}$, and $10^{5}$. As shown in Figure \ref{figure2}, the appropriate $\alpha_g$ values for the ResNet-56 and VGG-19 models are $10^{5}$ and $10^{2}$, respectively. This analysis reveals that models with higher FLOPs require a smaller gravity compared to those with lower FLOPs to achieve effective pruning.

\section{Conclusion}
We introduced a novel method that integrates the concept of gravity from physics into DCNNs training through regularization. By applying gravitational forces to convolution filters, we either attract or repel their weights toward non-zero or zero values, respectively, resulting in a sparser model. Filters farther from the attracting filter contribute to sparsity, while those closer retain density. This weight distribution enables effective pruning with minimal information loss. Our method demonstrated promising results on the CIFAR dataset, comparable to existing SOTA techniques. Future work will explore further applications of gravity-inspired approaches to model pruning.

\bibliographystyle{IEEEbib}
\bibliography{bibliography}
\end{document}